%% file: main.tex
\documentclass[runningheads]{llncs}
\usepackage{graphicx}

\usepackage{float}
\usepackage{multirow}
\usepackage{booktabs}
\usepackage{siunitx}

\begin{document}

\input{content/00A-coverpage}
\input{content/00B-abstract}

\input{content/01-introduction}
\input{content/02-related-work}
\input{content/03-1-method-data}
\input{content/03-2-models}

\input{content/04-experiments}
\input{content/05-user-study}

\input{content/06-discussion}

\newpage
\bibliographystyle{splncs04}
\bibliography{main}

\input{content/07A-supplementary}

\end{document}

%% file: content/00A-coverpage.tex
\title{Towards Ultrafast MRI via Extreme \textit{k}-Space Undersampling and Superresolution
}

\titlerunning{Towards Ultrafast MRI}
\author{Anonymized for review}
\author{Aleksandr Belov\inst{1, 2} \and
Joël Valentin Stadelmann\inst{1} \and
Sergey Kastryulin\inst{1, 2} \and
Dmitry V. Dylov\inst{2}}
\authorrunning{A. Belov et al.}
%
\institute{Philips Innovation Labs RUS, Moscow, Russian Federation 
\email{\{aleksandr.belov,joel.stadelmann,sergey.kastryulin\}@philips.com}
\and
Skolkovo Institute of Science and Technology, Moscow, Russian Federation
\email{d.dylov@skoltech.ru}}

\maketitle

%% file: content/00B-abstract.tex
\begin{abstract}
We went below the MRI acceleration factors (\textit{a.k.a.}, \textit{k}-space undersampling) reported by all published papers that reference the original fastMRI challenge \cite{fastmri_data}, and then considered powerful deep learning based image enhancement methods to compensate for the underresolved images. 
We thoroughly study the influence of the sampling patterns, the undersampling and the downscaling factors, as well as the recovery models on the final image quality for both the brain and the knee fastMRI benchmarks.
The quality of the reconstructed images surpasses that of the other methods, yielding an MSE of $11.4 \cdot 10^{-4}$, a PSNR of $29.6$ dB, and an SSIM of $0.956$ at $\times16$ acceleration factor.
More extreme undersampling factors of $\times32$ and $\times64$ are also investigated, holding promise for certain clinical applications such as computer-assisted surgery or radiation planning.
We survey $5$ expert radiologists to assess 100 pairs of images and show that the recovered undersampled images statistically preserve their diagnostic value.




\keywords{Fast MRI \and Superresolution \and Image-to-image translation.}
\end{abstract}

%% file: content/01-introduction.tex
\section{Introduction}

Magnetic Resonance Imaging (MRI) is a non-invasive imaging modality that offers excellent soft tissue contrast and does not expose the patient to ionizing radiation. 
The raw MRI signal is recorded in the so-called \textit{k}-space, with the digital images being then generated by Fourier transform.
The filling of the \textit{k}-space typically lasts 15--60 minutes, during which the patient must remain motionless (problematic for children, neurotic or claustrophobic patients).
If a patient cannot stay still, a motion artifact will appear on the images, oftentimes demanding a complete re-scan \cite{card_artefacts}.
Furthermore, the long acquisition time limits the applicability of MRI to dynamic imaging of the abdomen or the heart \cite{rapid_compressed_sens,compressed_sens} and decreases the throughput of the scanner, leading to higher costs \cite{ultra_fast}.

Contrasting chemicals \cite{DebatinUltrafastMRIBook} and physics-based acceleration approaches \cite{PrakkamakulUltrafast,TsaoUltrafastImaging} alleviate the challenge and have been a subject of active research over the last two decades, solidifying the vision of the `Ultrafast MRI' as the ultimate goal.
A parallel pursuit towards the same vision is related to compressed sensing \cite{compressed_sens,rapid_compressed_sens,DebatinUltrafastMRIBook}, where the methods to compensate for the \textit{undersampled} \textit{k}-space by proper image reconstruction have been proposed. This direction of research experienced a noticeable resurgence following the publication of the \texttt{fastMRI} benchmarks \cite{fastmri_data} and the proposal to remove the artifacts resulting from the gaps in the \textit{k}-space by deep learning (DL) \cite{fastmri_challenge}. Given the original settings defined by the challenge, the majority of groups have been experimenting with the reconstruction for the acceleration factors of $\times$2, $\times$4, and $\times$8, with only rare works considering stronger undersampling of $\times$16 and $\times$32.

In this article, we further unholster the arsenal of DL and propose the use of powerful image-to-image translation models, originally developed for natural images, to tackle the poor quality of the underresolved MRI reconstructions. Specifically, we retrain Pix2Pix \cite{pix2pix} and SRGAN \cite{srgan} on the \texttt{fastMRI} data and study their performance alongside the reconstructing U-Net \cite{unet} at various downscaling and undersampling factors and sampling patterns. \emph{Our best model outperforms state-of-the-art (SOTA) at all conventional acceleration factors and allows to go beyond them to attempt extreme $k$-space undersampling, such as $\times64$.}

%% file: content/02-related-work.tex
\section{Related work.}
Several DL approaches have already been implemented to accelerate MRI. 
For instance,
\cite{joint} proposed an algorithm that combines the optimal cartesian undersampling and MRI reconstruction using U-Net \cite{unet}, which increased the fixed mask PSNR by $0.45 - 0.8$ dB and SSIM by $0.012 - 0.017$.
The automated transform by manifold approximation (AUTOMAP)
\cite{automap} learns a transform between \textit{k}-space and the image domain using fully-connected layers and convolutions layers in the image domain. The main disadvantage of AUTOMAP is a quadratic growth of parameters with the number of pixels in the image.
Ref. \cite{k-sp} presented an approach, based on the interpolation of the missing k-space data using low-rank Hankel matrix completion, that consistently outperforms existing image-domain DL approaches on several masks. 
\cite{philips_fastmri} used a cascade of U-Net modules to mimic iterative image reconstruction, resulting in an SSIM of $0.928$ and a PSRN of $39.9$ dB at $\times8$ acceleration on multi-coil data. As the winner of the challenge, this model is considered SOTA.

%
%
%
%
%
%

The most recent works tackle the same challenge by applying parallel imaging \cite{brain_mri_time}, generative adversarial networks \cite{chen2020mri}, ensemble learning with priors \cite{Lyu_2020}, trajectory optimization \cite{wang2021bspline}, and greedy policy search \cite{BakkerHW20} (the only paper we found that considered the factor $\times32$). The issues of DL-based reconstruction robustness and the ultimate clinical value have been outlined in multiple works (\textit{e.g.}, \cite{darestani2021measuring,wang2021bspline}), raising a reasonable concern of whether the images recovered by DL preserve their diagnostic value. Henceforth, we decided to engage a team of radiologists to perform such a user study to review our results.

%% file: content/03-1-method-data.tex
\section{Method}
\subsection{Data description}

The Brain \texttt{fastMRI} dataset contains around \num[group-separator={,}]{10000} anonymized MRI volumes provided by the NYU School of Medicine \cite{fastmri_data}. We chose only T2 scans (as the largest class), padded, and downsized them to $320\times320$ pixels, discarding the smaller images.
Finally, we selected $9$ slices of each MRI volume at mid-brain height to obtain similar brain surface area in each image.
The resulting \num[group-separator={,}]{73478} slices were split into train\hspace{0.05cm}/\hspace{0.05cm}validation\hspace{0.05cm}/\hspace{0.05cm}test subsets as 60\%\hspace{0.05cm}/\hspace{0.05cm}20\%\hspace{0.05cm}/\hspace{0.05cm}20\%, making sure that slices from a given patient belong within the same subset.
We applied min-max scaling of the intensities using the 2\textsuperscript{nd} and the 98\textsuperscript{th} percentiles.

We also validated our methods on the MRI scans of the knee. 
The same data preparation resulted in \num[group-separator={,}]{167530} T2 slices, split into the 
subsets by the same proportion. 
Synovial liquid \cite{DebatinUltrafastMRIBook} noticeably shifted the upper percentiles in the knee data; thus, we normalized the intensities using the distance between the first and the second histogram extrema instead. 

%% file: content/03-2-models.tex
\subsection{Models}

\begin{figure}[b]
\begin{center}
\vskip -0.2in
\centerline{\includegraphics[width=\textwidth]{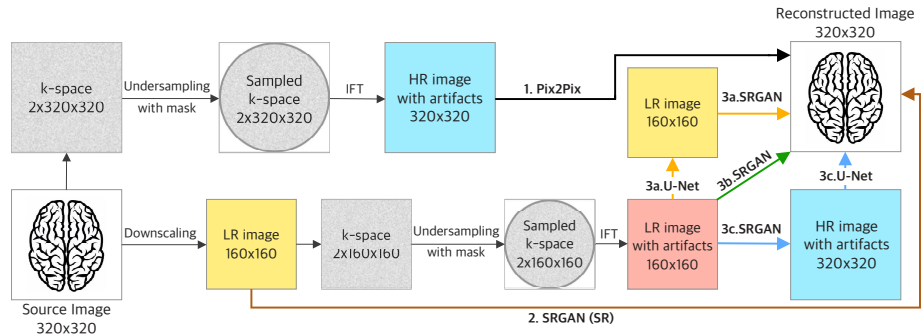}}
\caption{Three possible approaches to downscale and undersample $k$-space. The sampling masks details are given in the supplementary material.}
\label{all_models}
\end{center}
\vskip -0.2in
\end{figure}

An overview of the different models studied herein is shown in Fig. \ref{all_models}. Using the compressed sensing paradigm \cite{DebatinUltrafastMRIBook,fastmri_challenge,compressed_sens}, we aim to accelerate MRI 
by sampling low resolution or/and partially filled \textit{k}-space.
The partial filling is accomplished by applying patterned masks on the Fourier transform of the original slices.
We used the masks proposed by the \texttt{fastMRI} challenge (a central vertical band of $8\%$ of the value is fixed and the remaining bands are sampled randomly), as well as the \textit{radial} and the \textit{spiral} sampling trajectories that are popular in the field of compressed sampling \cite{sampling_traj}.
The three types of masks considered and the effect of them on the images are shown in the supplementary material.

We used Pix2Pix \cite{pix2pix} as an image-to-image translation method to correct the artifacts introduced by the \textit{k}-space undersampling and SRGAN \cite{srgan} to upscale the low-resolution images to their original size of $320\times320$. 
Interested in preserving the high-frequency components, we used the MSE of the feature maps obtained with the VGG16 pre-trained on \texttt{ImageNet} 31\textsuperscript{th} layer for the perceptual loss, relying on the reports that convolutional networks pre-trained on \texttt{ImageNet} are still relevant for the perceptual losses of MRI images \cite{johnson2016perceptual,percep_loss_mri,synth_ct_mri}.

We retrained Pix2Pix and SRGAN (SR, for superresolution alone) entirely on the datasets described above. The combinations of SRGAN and U-Net generators (paths 3a-c in Fig. \ref{all_models}) were trained with SRGAN discriminator and a joint back-propagation. We also considered SRGAN for the direct reconstruction without the U-Net (path 2 in Fig. \ref{all_models}).
The quality of the produced images was assessed using the MSE, PSNR and SSIM \cite{ssim} metrics, similarly to SOTA \cite{fastmri_challenge,philips_fastmri}. 

%% file: content/04-experiments.tex
\section{Experiments}

We ran extensive series of experiments outlined in Fig. \ref{all_models}.
In all SRGAN variants, our models always outperformed bicubic interpolation, which we confirmed for the $\times2$, $\times4$, and $\times8$ upscaling factors.
We also observed that Pix2Pix models, following the undersampling with the radial and the spiral masks, always outperformed the \texttt{fastMRI} mask. The difference between the radial and the spiral masks, however, was less pronounced (see Tables \ref{pix2pix_results} and \ref{sr_results}). 
All experimental results are shown in the format of $\mu \pm \sigma$, where $\mu$ is the average metric value over the test set, and $\sigma$ is its standard deviation.

\vskip -0.4in

\begin{figure}[ht]
\begin{center}

\hspace{-0.6cm}
\begin{minipage}[h]{0.45\linewidth}

\begin{table}[H]
\scalebox{0.55}{

\begin{tabular}{|p{37pt}|p{37pt}|p{62pt}|p{62pt}|p{62pt}|}
\hline
Acc. /   & \multirow{3}{4em}{Mask} & \multirow{3}{8em}{MSE, $\times10^{-4}$}  & \multirow{3}{4em}{PSNR} & \multirow{3}{8em}{SSIM, $\times 10^{2}$}   \\ 
frac. of &&&& \\
\textit{k}-space &&&& \\
\hline\hline
\multirow{3}{4em}{$\times2$ \\ 50\% }  & fastMRI  & 14.39 $\pm$ 3.33  &  28.53 $\pm$ 0.96  &  94.02 $\pm$ 3.19 \\ 
                                       & spiral   & 6.14  $\pm$ 2.13  &  32.37 $\pm$ 1.47  &  96.84 $\pm$ 2.22 \\                            
                                       & \textbf{radial}    & \textbf{5.08  $\pm$ 1.83}  &  \textbf{33.20 $\pm$ 1.47}  &  \textbf{97.70 $\pm$ 1.74} \\
\hline
\multirow{3}{4em}{$\times4$ \\ 25\%}   & fastMRI  & 27.93 $\pm$ 6.07  &  25.64 $\pm$ 0.93  &  90.65 $\pm$ 3.95 \\
                                       & spiral   & 13.96 $\pm$ 3.81  &  28.70 $\pm$ 1.15  &  93.56 $\pm$ 3.52 \\ 
                                       & \textbf{radial}    & \textbf{11.81 $\pm$ 4.62}  &  \textbf{29.51 $\pm$ 1.37}  &  \textbf{94.46 $\pm$ 3.93} \\
\hline
\multirow{3}{4em}{$\times8$ \\ 12.5\%} & fastMRI  & 54.41 $\pm$ 12.13 &  22.75 $\pm$ 0.98  &  85.74 $\pm$ 4.21 \\ 
                                       & spiral   & 28.00 $\pm$ 7.67  &  25.68 $\pm$ 1.13  &  89.45 $\pm$ 4.38 \\ 
                                       & \textbf{radial}    & \textbf{21.12 $\pm$ 5.45}  &  \textbf{26.89 $\pm$ 1.08}  &  \textbf{91.23 $\pm$ 4.61} \\
\hline
\multirow{2}{4em}{$\times16$ \\ 6.25\%} & spiral  & 45.04 $\pm$ 9.89  &  23.57 $\pm$ 1.00  &  85.93 $\pm$ 3.90 \\ 
                                        & \textbf{radial}   & \textbf{42.21 $\pm$ 9.78}  &  \textbf{23.87 $\pm$ 1.07}  &  \textbf{87.12 $\pm$ 4.53} \\ 
\hline
\multirow{2}{4em}{$\times32$ \\ 3.125\%} & spiral  & 69.00 $\pm$ 17.48  &  21.76 $\pm$ 1.17  &  82.37 $\pm$ 4.84 \\ 
                                         & \textbf{radial}  & \textbf{65.62 $\pm$ 19.70}  &  \textbf{22.02 $\pm$ 1.31}  &  \textbf{83.06 $\pm$ 5.76} \\ 
\hline
\end{tabular}
}
\vskip 0.1in
\caption{Pix2Pix model results.}
\label{pix2pix_results}
\end{table}

\end{minipage}
\hspace{0.05cm}
\begin{minipage}[h]{0.45\textwidth}%

\begin{table}[H]
\raisebox{-2.8cm}{
\scalebox{0.55}{
\begin{tabular}{|p{37pt}|p{41pt}|p{63pt}|p{63pt}|p{63pt}|}
\hline
Acc. /   & \multirow{3}{4em}{Models} & \multirow{3}{8em}{MSE, $\times10^{-4}$}  & \multirow{3}{4em}{PSNR} & \multirow{3}{8em}{SSIM, $\times 10^{2}$}   \\ 
frac. of &&&& \\
\textit{k}-space &&&& \\
\hline\hline
\multirow{2}{4em}{$\times4$ \\ 25\% }     & Bicubic          & 10.51 $\pm$ 4.29   &  30.16 $\pm$ 1.84  &  96.85 $\pm$ 1.51 \\ 
                                          & \textbf{SRGAN}   & \textbf{3.03  $\pm$ 1.61}   &  \textbf{35.72 $\pm$ 2.12}  &  \textbf{98.64 $\pm$ 0.93} \\                            
\hline 
\multirow{2}{4em}{$\times16$ \\ 6.25\%}   & Bicubic          & 43.94 $\pm$ 14.66  &  23.84 $\pm$ 1.58  &  88.48 $\pm$ 3.17 \\
                                          & \textbf{SRGAN}   & \textbf{11.69 $\pm$ 4.21}   &  \textbf{29.58 $\pm$ 1.49}  &  \textbf{95.63 $\pm$ 2.35} \\ 
\hline
\multirow{2}{4em}{$\times64$ \\ 1.5625\%} & Bicubic          & 121.59     $\pm$ 29.76 &  19.30 $\pm$ 1.21  &  73.04 $\pm$ 4.17 \\ 
                                          & \textbf{SRGAN}   & \textbf{28.76 $\pm$ 9.67}   &  \textbf{24.37 $\pm$ 1.13}  &  \textbf{88.49 $\pm$ 3.32} \\ 
\hline
\end{tabular}
}}
\vskip 0.1in
\caption{SRGAN (SR) model results.}
\label{sr_results}
\end{table}
\end{minipage}
\end{center}
\end{figure}

\vskip -0.5in

\begin{figure}[b]
    \begin{center}
    \centerline{\includegraphics[width=\textwidth]{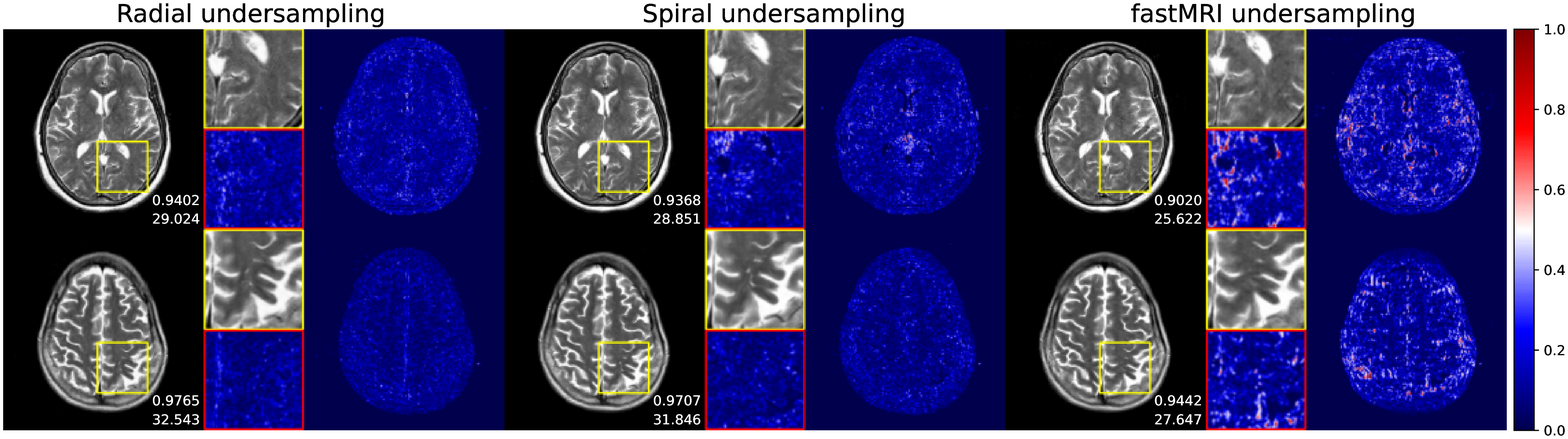}}
    \caption{Pix2Pix recovery results for $\times4$ \textit{k}-space undersampling. The difference with the ground truth (multiplied by 3 for clarity) are shown on the right. Shown in the corner are SSIM and PSNR values. Note the inferior performance of original fastMRI masks.}
    \label{pix2pix_x4}
    \end{center}
    \vskip -0.3in
\end{figure}

Figure \ref{pix2pix_x4} compares the Pix2Pix reconstruction at $\times4$ undersampling for different sampling patterns. Visually, the reconstructions of all models look clear with sharp anatomic structures. However, the difference images reveal that the finest details are reconstructed best via the radial sampling. This naturally stems from the way the radial mask samples the highest spatial frequencies.

\begin{figure}[H]
    \begin{center}
    \centerline{\includegraphics[width=\textwidth]{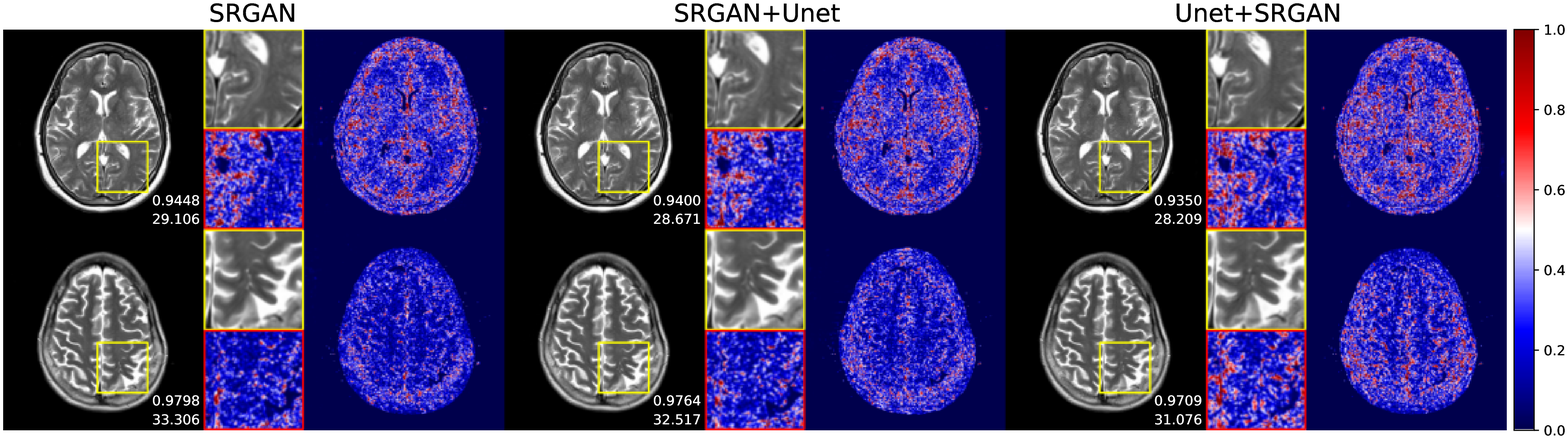}}
    \caption{Recovery of $\times8$ Low-Resolution \textit{k}-space (undersampling with the radial mask). The subtraction from the ground truth (multiplied by 3 for clarity and shown on the right) suggests the SRGAN+U-Net combo to be the best model for MRI acceleration. 
    }
    \label{srgan_unet_x8}
    \end{center}
    \vskip -0.2in
\end{figure}

For the combination of architectures, we downscaled the images to $160\times160$ pixels before applying \textit{k}-space undersampling. We then trained SRGAN for both the upscaling and artifact removal or SRGAN+U-Net / U-Net+SRGAN, depending on the order of application of the corresponding generators. Visually, SRGAN alone performed worse than SRGAN+U-Net and U-Net+SRGAN combos, both of which seem to be statistically equivalent (see Tables \ref{x8_combined_results}, \ref{x16_combined_results}, and Figure \ref{srgan_unet_x8}). A comprehensive presentation of the results obtained at different acceleration factors can be found in the supplementary material.

\begin{figure}[ht]
\vskip -0.4in
\begin{center}

\hspace{-0.85cm}
\begin{minipage}[h]{0.44\linewidth}

\begin{table}[H]
\scalebox{0.53}{
\begin{tabular}{|p{35pt}|p{76pt}|p{63pt}|p{63pt}|p{63pt}|}
\hline
Mask                       & Model        & MSE, $\times10^{-4}$ & PSNR            & SSIM, $\times 10^{2}$  \\ 
\hline\hline
\multirow{3}{4em}{fastMRI} & SRGAN        & 12.98 $\pm$ 3.79  &  29.04 $\pm$ 1.22  &  94.84 $\pm$ 2.55 \\ 
                           & SRGAN+U-Net   & 8.81  $\pm$ 3.08  &  30.80 $\pm$ 1.44  &  95.97 $\pm$ 2.17 \\ 
                           & U-Net+SRGAN   & 9.27  $\pm$ 3.28  &  30.56 $\pm$ 1.41  &  95.69 $\pm$ 2.69 \\
\hline
\multirow{3}{4em}{spiral}  & SRGAN        & 7.66  $\pm$ 2.94  &  31.47 $\pm$ 1.66  &  96.28 $\pm$ 2.26 \\
                           & SRGAN+U-Net   & 7.02  $\pm$ 2.84  &  31.88 $\pm$ 1.73  &  96.51 $\pm$ 2.06 \\ 
                           & U-Net+SRGAN   & 7.09  $\pm$ 2.82  &  31.81 $\pm$ 1.65  &  96.39 $\pm$ 2.37 \\
\hline
\multirow{3}{4em}{radial}  & SRGAN        & 6.25  $\pm$ 2.63  &  32.40 $\pm$ 1.78  &  96.95 $\pm$ 1.95 \\ 
                           & \textbf{SRGAN+U-Net}   & \textbf{6.05  $\pm$ 2.64}  &  \textbf{32.58 $\pm$ 1.87}  &  \textbf{97.01 $\pm$ 1.87} \\ 
                           & U-Net+SRGAN   & 6.13  $\pm$ 2.62  &  32.48 $\pm$ 1.75  &  96.92 $\pm$ 2.11 \\
\hline
\end{tabular}
}
\vskip 0.1in
\caption{Low-Resolution ($160 \times 160$) model with $\times2$ undersampling, total acceleration: $\times8$, fraction of \textit{k}-space: $12.5\%$}

\label{x8_combined_results}
\end{table}

\end{minipage}
\hspace{0.5cm}
\begin{minipage}[h]{0.44\textwidth}%

\begin{table}[H]
\scalebox{0.53}{
\begin{tabular}{|p{35pt}|p{76pt}|p{63pt}|p{63pt}|p{63pt}|}
\hline
Mask                       & Model        & MSE, $\times10^{-4}$ & PSNR            & SSIM, $\times 10^{2}$  \\ 
\hline\hline
\multirow{3}{4em}{fastMRI} & SRGAN        & 23.07 $\pm$ 6.30  &  26.53 $\pm$ 1.17  &  92.40 $\pm$ 2.93 \\ 
                           & SRGAN+U-Net   & 15.97 $\pm$ 4.77  &  28.15 $\pm$ 1.26  &  94.00 $\pm$ 2.61 \\ 
                           & U-Net+SRGAN   & 18.55 $\pm$ 5.88  &  27.51 $\pm$ 1.28  &  93.32 $\pm$ 3.27 \\
\hline
\multirow{3}{4em}{spiral}  & SRGAN        & 17.29 $\pm$ 5.05  &  27.79 $\pm$ 1.20  &  93.32 $\pm$ 2.88 \\
                           & SRGAN+U-Net   & 13.61 $\pm$ 4.11  &  28.85 $\pm$ 1.26  &  94.19 $\pm$ 2.58 \\ 
                           & U-Net+SRGAN   & 13.49 $\pm$ 4.18  &  28.89 $\pm$ 1.27  &  94.08 $\pm$ 3.39 \\
\hline
\multirow{3}{4em}{radial}  & SRGAN        & 12.71 $\pm$ 4.07  &  29.18 $\pm$ 1.38  &  94.54 $\pm$ 2.79 \\ 
                           & \textbf{SRGAN+U-Net}   & 11.44 $\pm$ 3.90  &  29.66 $\pm$ 1.46  &  \textbf{94.92 $\pm$ 2.59} \\ 
                           & U-Net+SRGAN   & \textbf{11.39 $\pm$ 3.92}  &  \textbf{29.67 $\pm$ 1.42}  &  94.79 $\pm$ 3.23 \\
\hline
\end{tabular}
}
\vskip 0.1in
\caption{ Low-Resolution ($160 \times 160$) model with $\times4$ undersampling, total acceleration: $\times16$, fraction of \textit{k}-space: $6.25\%$}
\label{x16_combined_results}
\end{table}

\end{minipage}
\end{center}
\end{figure}

\vskip -0.6in

\begin{figure}[ht]
\vskip -0.5in
\begin{center}

\begin{minipage}[h]{0.47\linewidth}

\centerline{\includegraphics[width=1.2\textwidth]{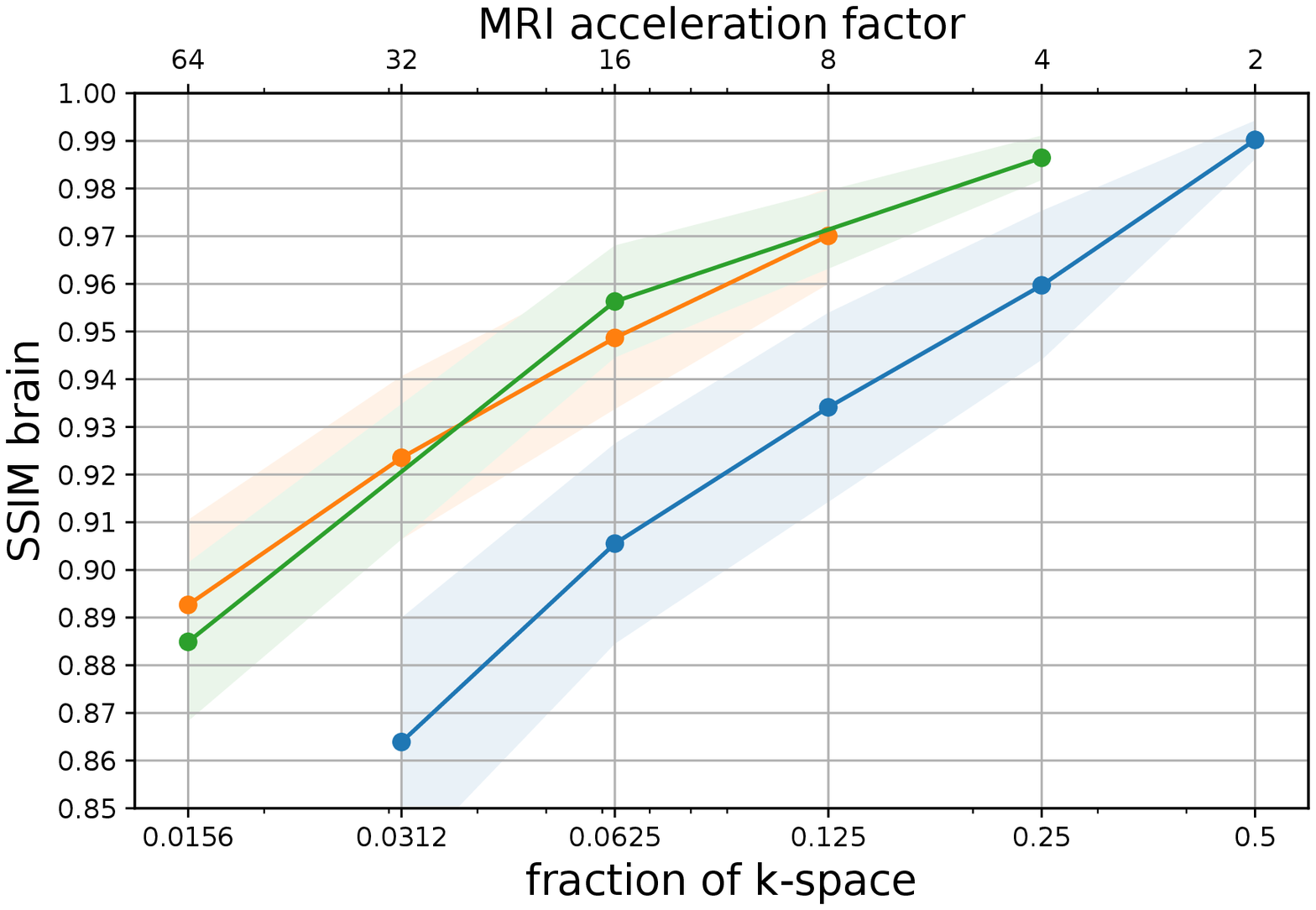}}
\caption{SSIM dependence on the fraction of \textit{k}-space for several best models, the brain dataset.}
\label{brain_ssim_comparison}

\end{minipage}
\hfill 
\begin{minipage}[h]{0.47\textwidth}

\centerline{\includegraphics[width=1.2\textwidth]{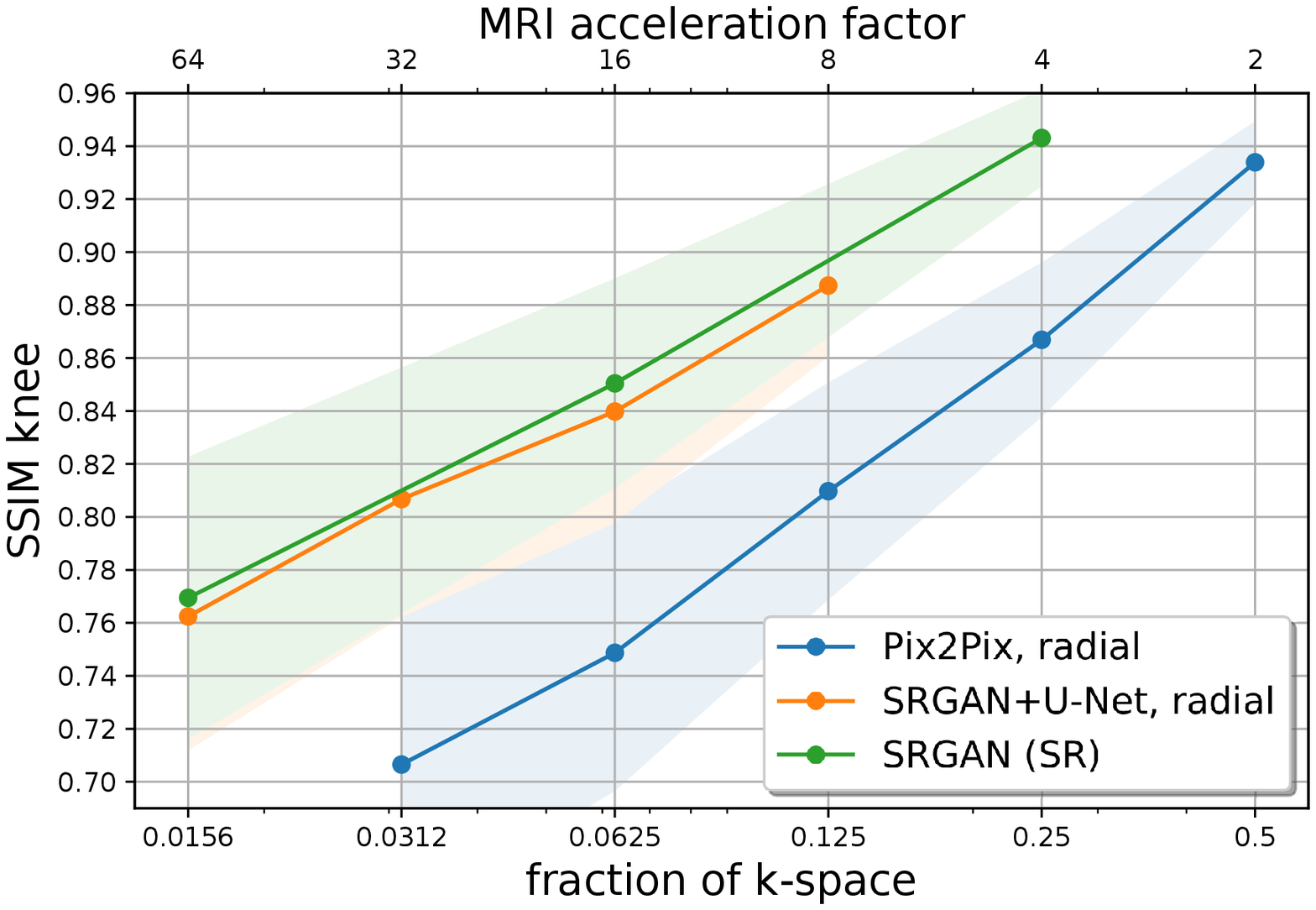}}
\caption{SSIM dependence on the fraction of \textit{k}-space for several best models, the knee dataset.}
\label{knee_ssim_comparison}

\end{minipage}
\end{center}

\vskip -0.3in
\end{figure}

Our experiments show that undersampling with the radial mask, both for the Pix2Pix models and for the SRGAN+U-Net combo, always outperforms the other masks. Interestingly, SRGAN (SR) yields slightly better scores than the combo methods up to the $\times16$ acceleration (see Figure \ref{brain_ssim_comparison}) despite the visual analysis where the combos are somewhat more convincing to the eye of an expert. 

\begin{figure}[ht]
\begin{center}

\begin{minipage}[h]{0.50\linewidth}

\centerline{\includegraphics[scale=0.36]{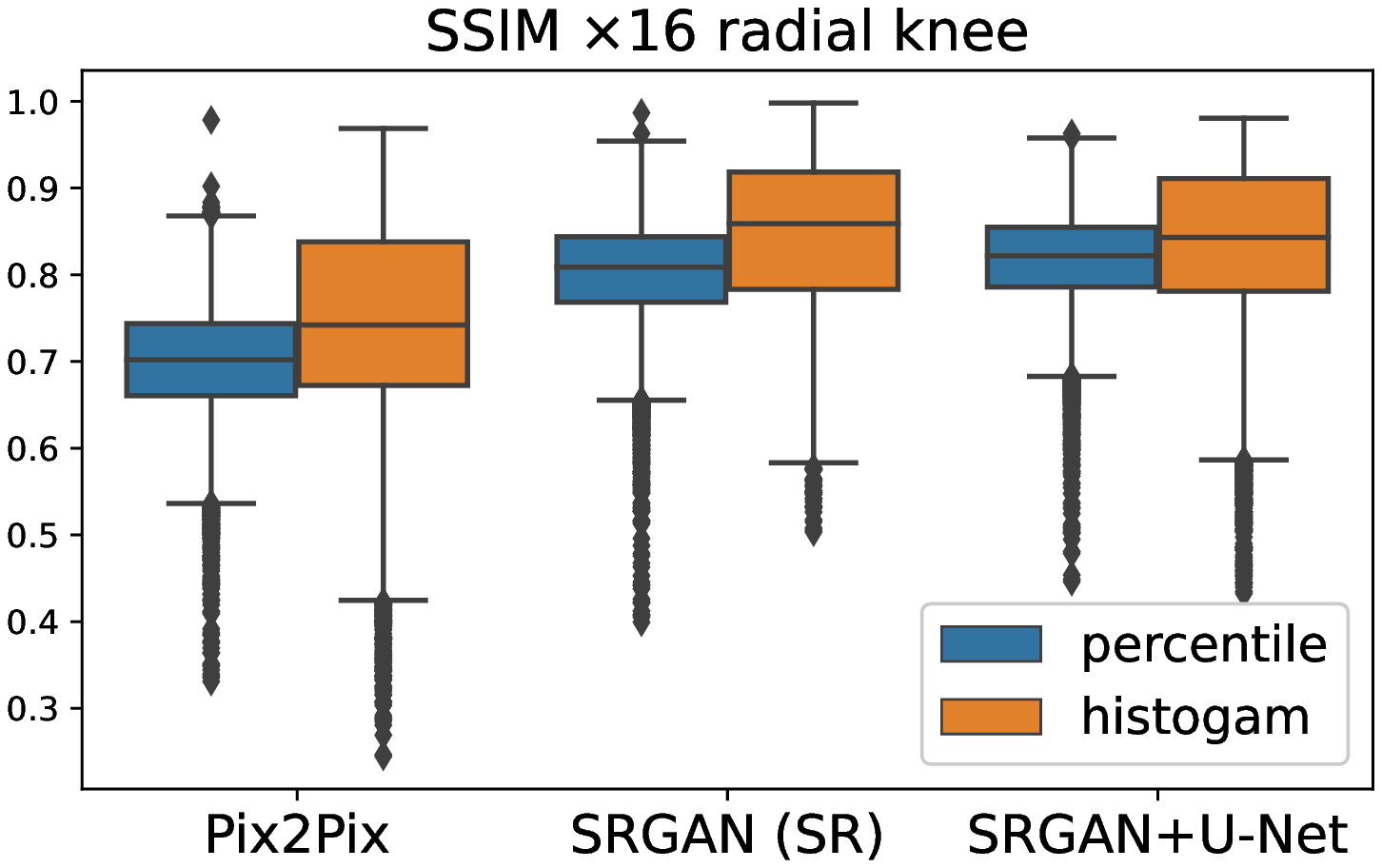}}
\caption{Influence of the normalization method on the SSIM for the knee dataset with $\times$16 acceleration and radial sampling.}
\label{ssim_16_knee}

\end{minipage}
\hspace{0.1cm}
\begin{minipage}[h]{0.47\textwidth}

\raisebox{-0.05cm}{
\centerline{\includegraphics[scale=0.21]{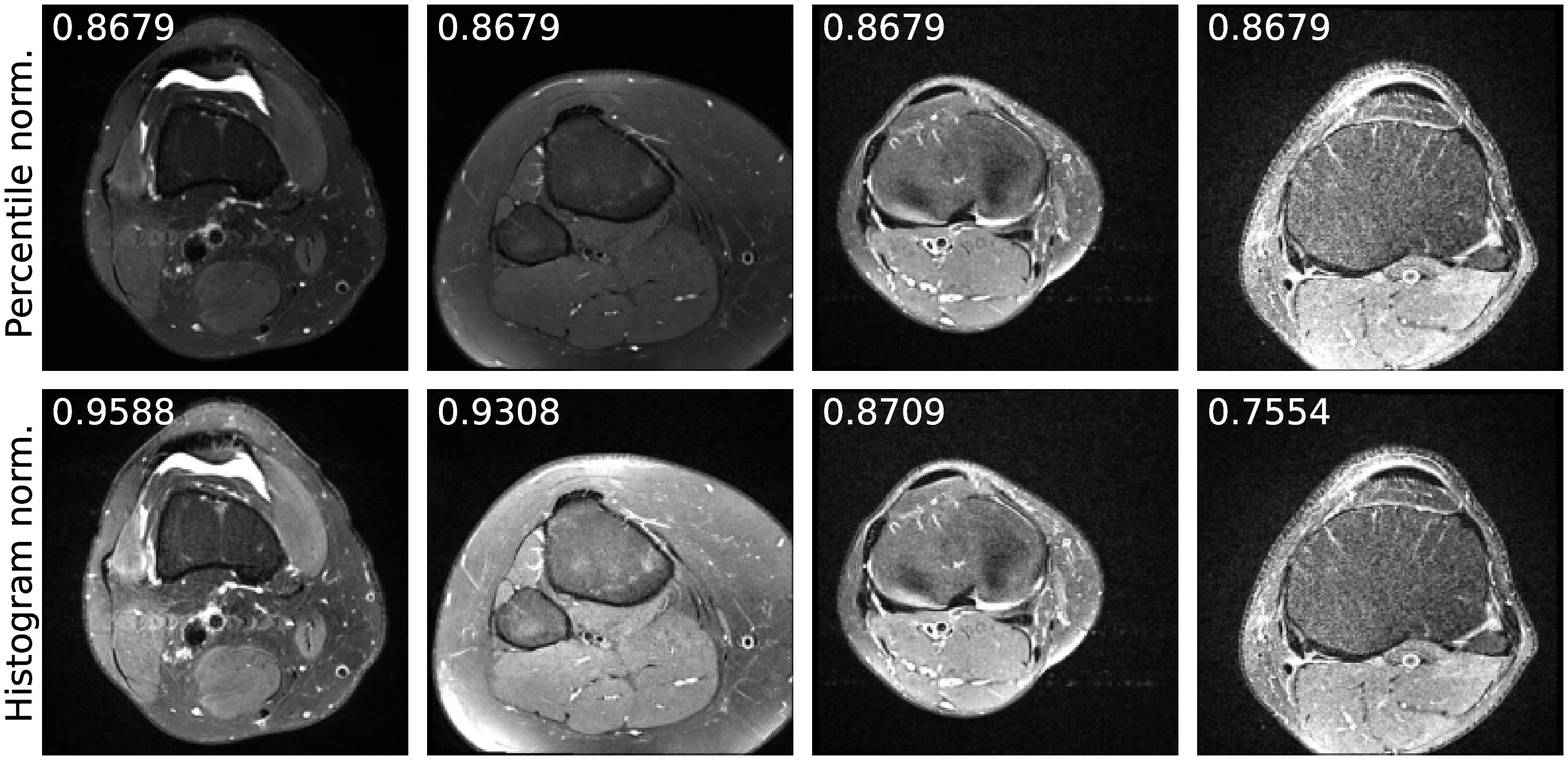}}
}
\caption{Knee slice reconstructions with percentile-based (top) and histogram-based normalization (bottom).}
\label{knee_ex}

\end{minipage}
\end{center}
\end{figure}

We applied the same models to the knee dataset, which, at first, had led to a low image reconstruction quality (both visually and quantitatively). 
We hypothesized that the low quality of reconstruction could be caused by the differences in the pixel intensities after the normalization, with the same tissue type having been mapped to a different intensity range on two neighboring slices in the volume. Indeed, swapping percentile normalization with the histogram-based normalization yielded more homogeneous results of pixel intensities across the knee slices, which, consequently, boosted the quality of image reconstruction to the level of the results on the brain data (see Figure \ref{ssim_16_knee} and \ref{knee_ex}).




\vskip -0.2in

%% file: content/05-user-study.tex
\section{User study}
\label{sec:user-study}

Image quality (IQ) assessment metrics such as SSIM help to estimate the reconstruction quality.
However, they may not fully reflect how valuable these reconstructions are from the diagnostic perspective \cite{Esteban2019MRIIQM}.
To solve this problem, we conducted a user study of reconstruction quality with trained radiologists. 

The study was performed on a test subset of the fastMRI dataset. 
We randomly selected $25$ brain and $25$ knee slices. Those slices were downscaled with a $\times4$ factor and undersampled by factors of $\times2$ and $\times4$, using the radial masks. 
The resulting $100$ slices were evaluated for their diagnostic content, compared to their corresponding fully-sampled counterparts (the ground truth).
%

Five radiologists were involved in the study, using a simple survey tool \cite{label-studio}. 
The experts were shown the ground truth images and asked to rate the quality of our superresolved reconstructions based on three criteria vital for a typical diagnostic decision-making \cite{DebatinUltrafastMRIBook}: the presence of artifacts, the signal-to-noise ratio (SNR), and the contrast-to-noise ratio (CNR). 
Each criterion was scored on a 4-point scale as follows: fully acceptable (4), rather acceptable (3), weakly acceptable (2), not acceptable (1). 

\begin{figure}[H]
    \centering
    \includegraphics[scale=0.28]{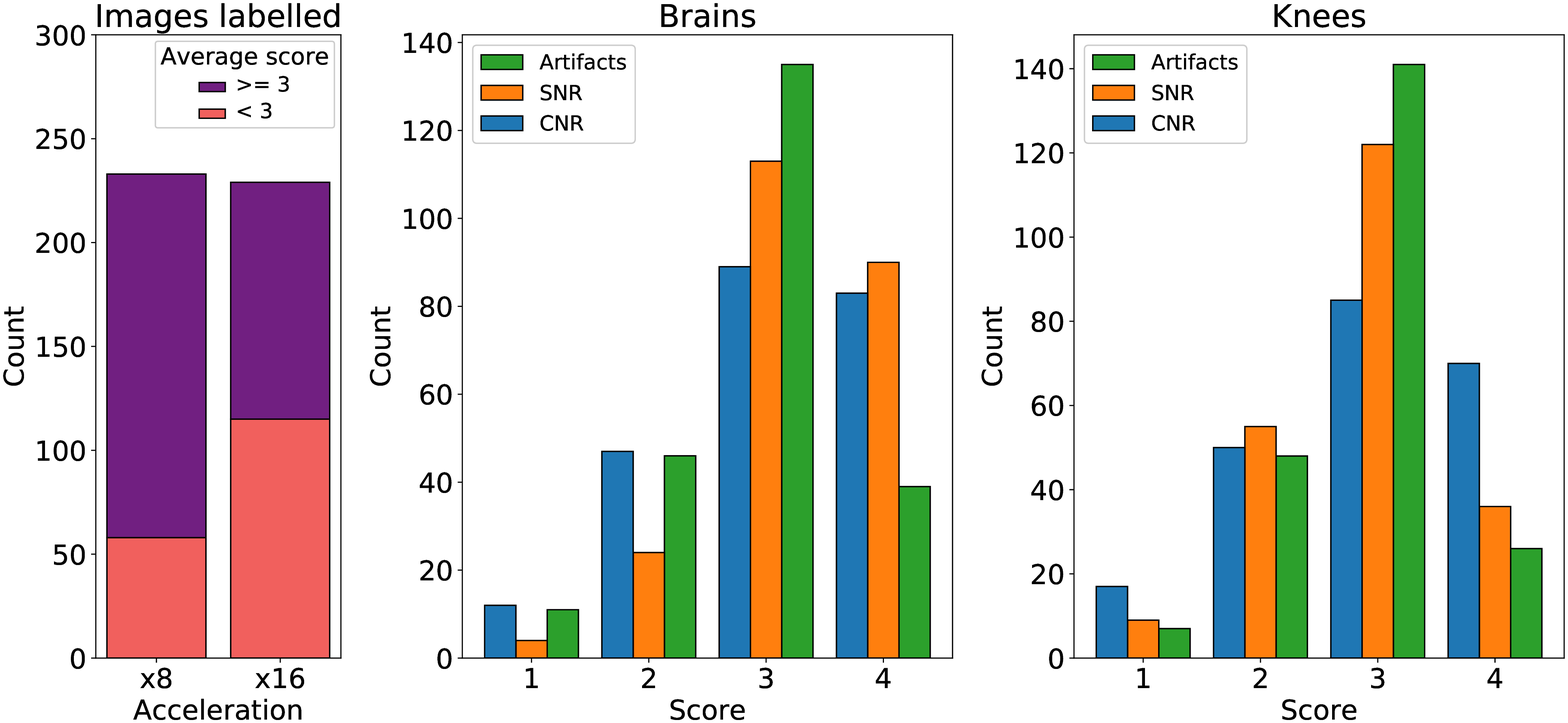}
    \caption{Diagnostic quality scores from a survey of radiologists based on three IQ parameters for brain and knee anatomies. The majority of votes fall into rather acceptable (3) and fully acceptable (4) categories, showing little discrepancy between anatomies.}
    \label{fig:user-study-results}
\end{figure}

Figure \ref{fig:user-study-results} shows high diagnostic value of the reconstructions, with around 80\% of images getting the scores of (3) and (4) for both anatomies.

%% file: content/06-discussion.tex
\section{Discussion and conclusion}

We considered powerful image-to-image translation and superresolution models to compensate for the strong undersampling and downscaling of the \textit{k}-space. 
Our study comprises a comprehensive evaluation of the undersampling patterns and factors, downscaling, and reconstruction models. 

Being more symmetrical and homogeneous at high frequencies, the radial and the spiral undersampling trajectories outperform corresponding original \texttt{fastMRI} sampling masks (for an equivalent fraction of \textit{k}-space).
Unlike them, the radial and the spiral sampling trajectories guarantee that the sampling is omnidirectional, without any given region of the \textit{k}-space being omitted, including the high-frequency ones. Following the radial or spiral sampling, the acquired data can be translated to the Cartesian grid according to the established apparatus in the compressed sensing \cite{compressed_sens}. It is natural, thus, to extrapolate our results to the other cutting-edge undersampling trajectories \cite{compressed_sens,ultra_fast,wang2021bspline}, which we anticipate to `heal' similarly well by the Pix2Pix model.

We found that until the $\times16$ acceleration factor, SRGAN (SR) marginally outperforms the SRGAN+U-Net combo, and both of these approaches noticeably outperform Pix2Pix. The absence of significant differences between SRGAN (SR), SRGAN+U-Net, and U-Net+SRGAN at a given fraction of \text{k}-space suggests that the latter two can be adopted in general as the go-to acceleration method, allowing for more flexibility in terms of computation efficiency and image quality trade-off (downscaled images require fewer parameters for the final reconstructing generator, which may influence the selection of one model \textit{vs.} the others for a particular application).

Different applications may indeed require a proper domain adaptation \cite{ghafoorian2017transfer}. Herein, \textit{e.g.}, the range of intensities in the knee data was noticeably skewed by the joint's liquids (the synovial fluid), requiring us to consider the histogram normalization approach in the data pre-processing stage (unlike percentile normalization sufficient for the brain data). More complex DL-based domain adaptation methods, translating the images between different MRI modalities and across different scanners \cite{ghafoorian2017transfer,synth_ct_mri}, can also be conjoined to the proposed MRI acceleration pipeline and will be the subject of future work. Quantitatively, with the histogram-based normalization, SRGAN (SR) yielded an MSE of $8.8 \cdot 10^{-4}$, a PSNR of $32.8$ dB, and an SSIM of $0.943$ at $\times4$ acceleration factor, which surpasses SOTA acceleration methods \cite{fastmri_challenge,philips_fastmri}.


Using the SRGAN+U-Net combo at $\times8$ acceleration, the reconstruction achieved an MSE of $6.1 \cdot 10^{-4}$, a PSNR of $32.6$ dB, and an SSIM of $0.97$. 
At $\times16$ acceleration, SRGAN+U-Net achieved an MSE of $11.4 \cdot 10^{-4}$, a PSNR of $29.6$ dB, and an SSIM of $0.956$ on the \texttt{fastMRI} brain data. 
These measurements correspond to a larger acceleration and a better ultimate image quality than those of the SOTA  methods \cite{philips_fastmri}.
The quantitative outcome is concordant with the results of our user study, where the differences between the brain and the knee images reflected those in the measured metrics in the radiologists' evaluation. 

In another ongoing survey, we are collecting the ranking votes for the pairs of images at the most extreme \textit{k}-space undersampling factors generated in our work ($\times32$ and $\times64$). While we do not anticipate those images to be useful for typical clinical examinations, the radiologists still express their enthusiasm because these extreme MRI acceleration factors would be of interest in certain other (non-diagnostic) applications, where the coarse images oftentimes prove sufficient (\textit{e.g.}, surgical planning in the radiological suites or radiation therapy planning in oncological care). We believe these extremely undersampled \textit{k}-space measurements are bound to improve with time due to the growth of the available data, which will gradually pave the way towards true ultrafast imaging of high diagnostic value.

%% file: content/07A-supplementary.tex
\newpage
\chapter*{\textsc{Supplementary material}}

\section*{A. Literature survey}
\begin{figure}[H]
\begin{center}
\centerline{\includegraphics[width=1.2\textwidth]{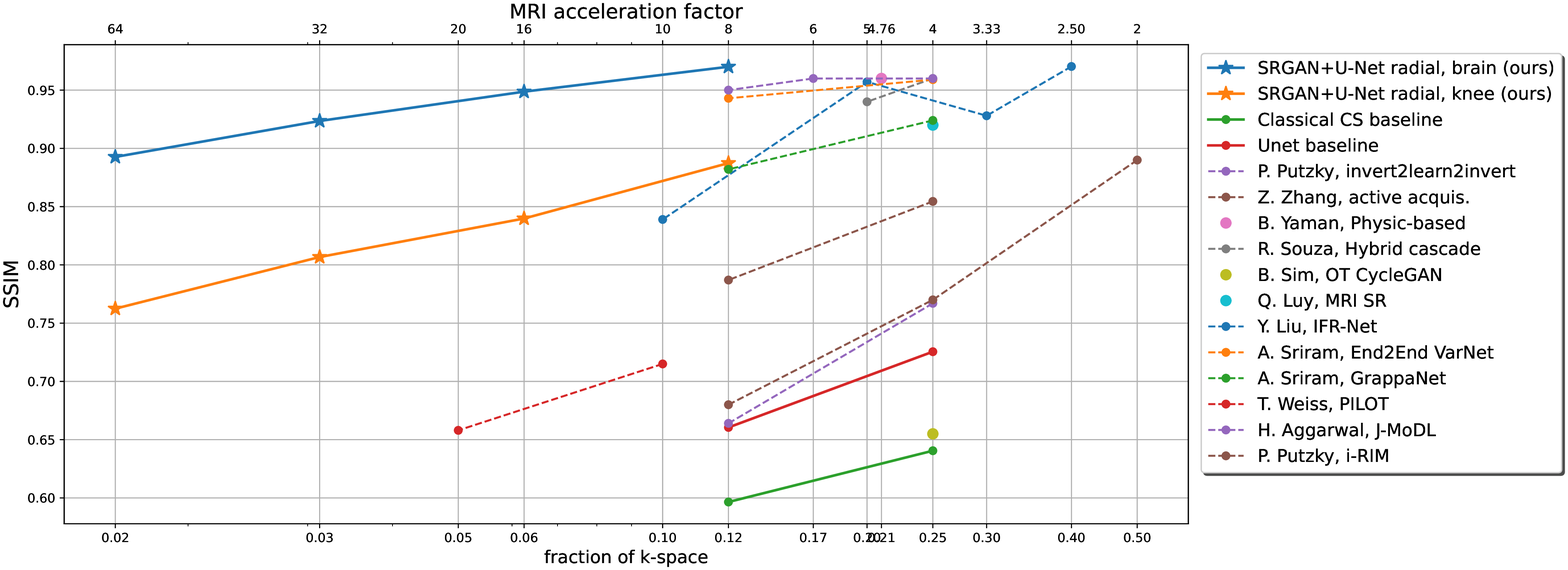}}
\caption{Comparison of acceleration factors and corresponding SSIM metrics of major papers that cite the original \texttt{fastMRI} dataset. Our models are in the top left corner.}
\label{related_work}
\end{center}
\end{figure}

\section*{B. Image intensity normalization}

We used two different normalization methods: percentile normalization and histogram-based normalization. 
\subsubsection*{Percentile normalization}
Percentile normalization is a min-max normalization to the $2^{nd}$ and $98^{th}$ percentiles of an image:
$$I_{norm} = \frac{I_{\text{source}} - p_{2\text{\%}}}{p_{98\text{\%}} - p_{2\text{\%}}}$$

Where $I_{\text{source}}$ is the intensity value of one pixel before normalization, $I_{\text{norm}}$ is its intensity value after normalization, $p_{2\text{\%}}$ and $p_{98\text{\%}}$, respectively are the $2^{nd}$ and $98^{th}$ percentiles of the intensity values for a given slice. After normalization, the values falling outside the $[0,1]$ range were clipped to those boundaries.
\clearpage
\subsubsection*{Histogram-based normalization}
Histogram-based normalization was performed as follows, for each image:
\begin{enumerate}
    \item Obtain the histogram of one slice;
    \item Fit a $15^{th}$ order polynomial to the histogram;
    \item Obtain the derivative of the polynomial;
    \item Search for zero-crossing of the derivative in order to locate the first minimum $m$ and maximum $M$ of the histogram;
    \item Center the new intensity range around $M$;
    \item Define the width of the intensity range as
    $$w = \alpha \cdot \Delta (M, m)$$
    where $\alpha = 5$ is a scaling coefficient, and $\Delta (M, m)$ is the distance between $M$ and $m$;
    \item Apply min-max normalization.
\end{enumerate}
The order of the polynomial, $15$, as well as the coefficient $\alpha=5$ were obtained empirically.
\section*{C. \textit{k}-space sampling trajectories}
\begin{figure}[H]
\begin{center}
\centerline{\includegraphics[width=0.6\textwidth]{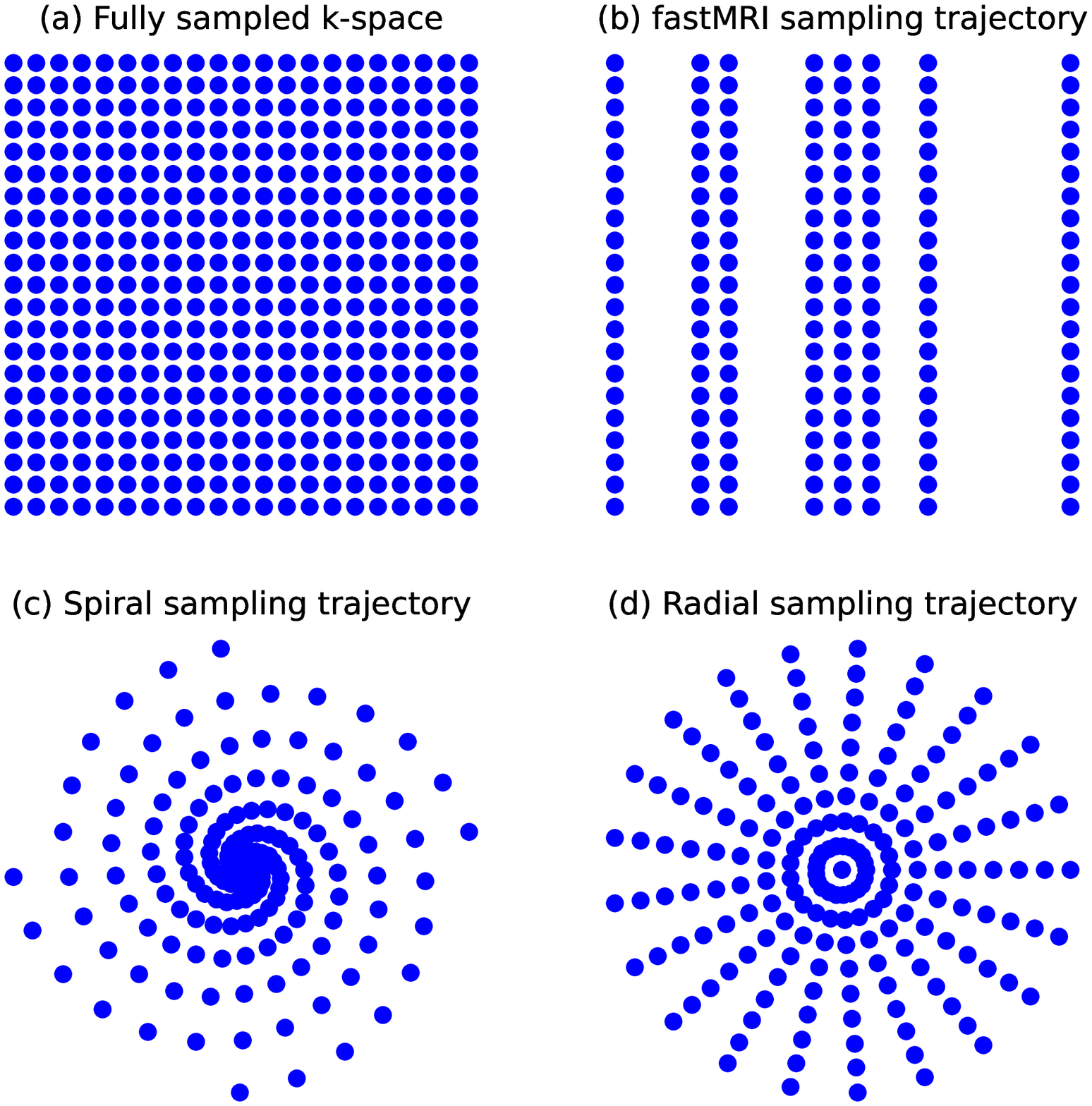}}
\caption{Typical patterns of the \textit{k}-space undersampling masks.}
\label{masks_scheme}
\end{center}
\vskip -0.2in
\end{figure}

\begin{figure}[H]
\begin{center}
\centerline{\includegraphics[width=\textwidth]{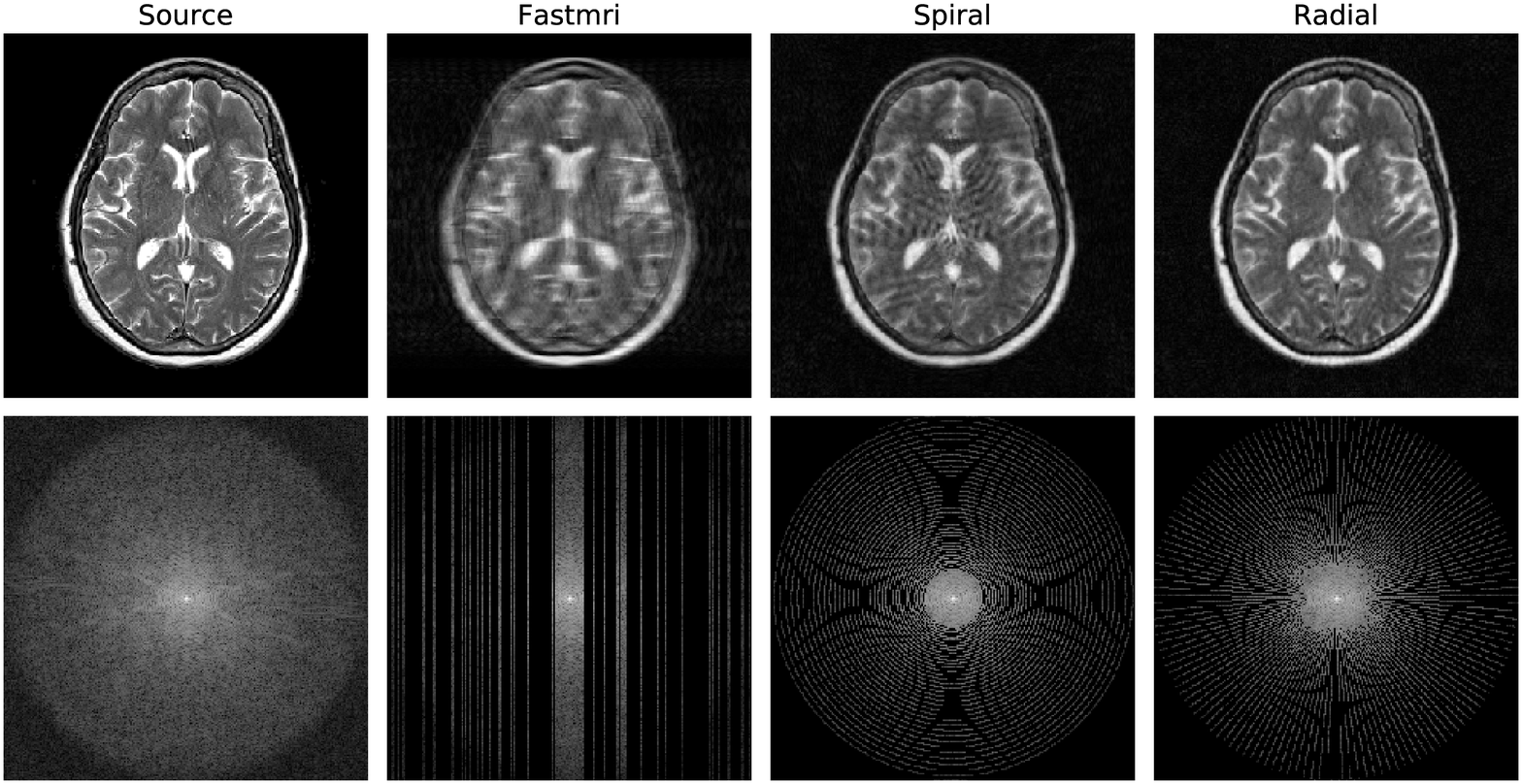}}
\caption{Examples of images obtained using a $\times4$ acceleration factor and different \textit{k}-space undersampling masks.}
\label{masks}
\end{center}
\end{figure}

\begin{figure}[ht]
\begin{center}

\begin{minipage}[h]{1\linewidth}

\centerline{\includegraphics[width=\textwidth]{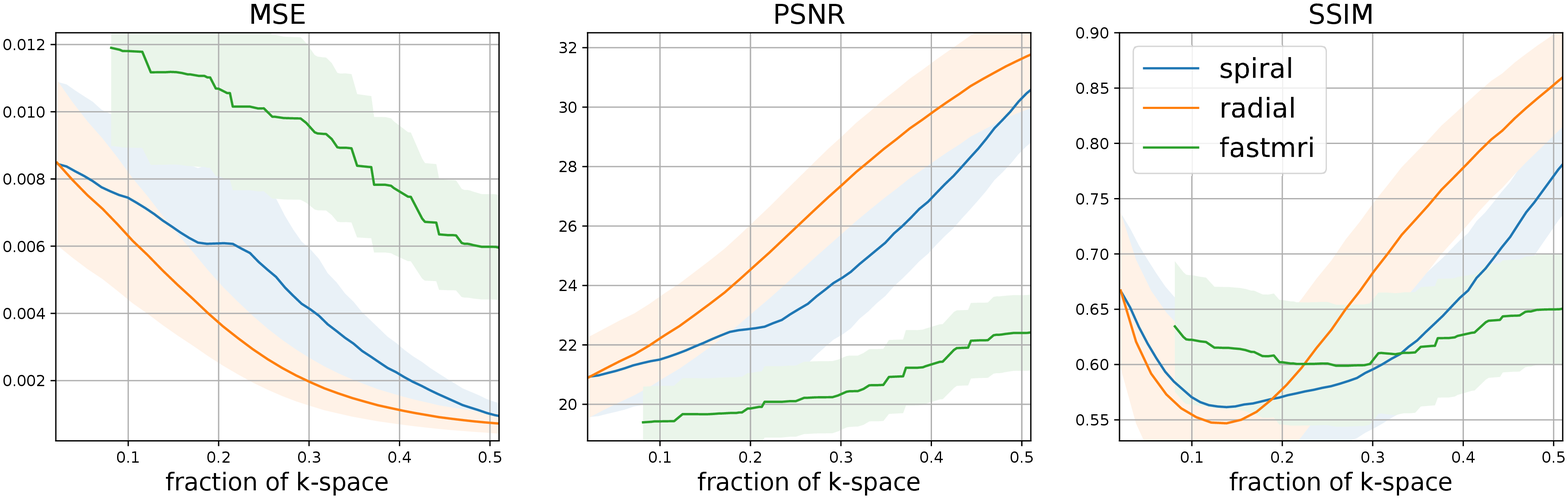}}
\centerline{(a) The brain dataset}
\label{brain_share}

\end{minipage}
\vfill 
\begin{minipage}[h]{1\textwidth}

\centerline{\includegraphics[width=\textwidth]{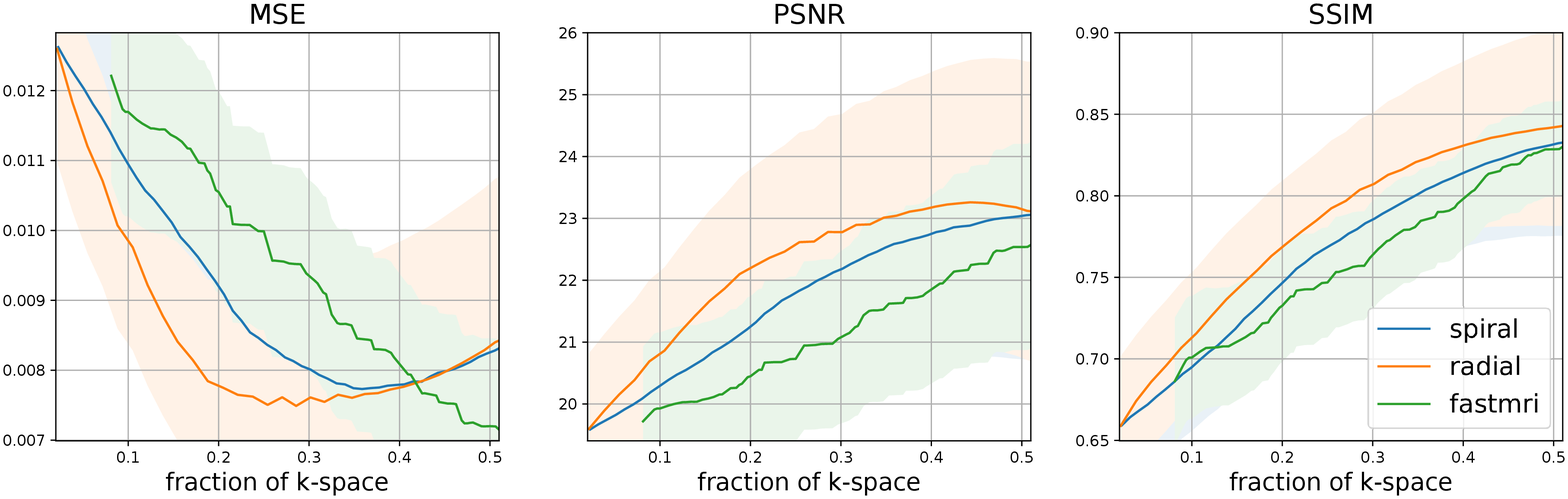}}
\centerline{(b) The knee dataset}
\label{knee_share}

\end{minipage}
\end{center}
\vskip -0.1in
\caption{Dependence of the target metrics on the \textit{k}-space fraction, calculated for the pairs of the ground truth and the undersampled images.}
\label{all_share}
\end{figure}

We present here the four trajectories used for the sampling of the \textit{k}-space (Figure \ref{masks_scheme}) as well as examples of images obtained with those trajectories (Figure \ref{masks}). Interestingly, without the reconstruction, \texttt{fastMRI} mask starts off with a marginally better image quality at low \textit{k}-space sampling (Figure \ref{all_share}); however, it quickly yields worse quality following the reconstruction.

\begin{figure}[H]
\begin{center}
\centerline{\includegraphics[width=\textwidth]{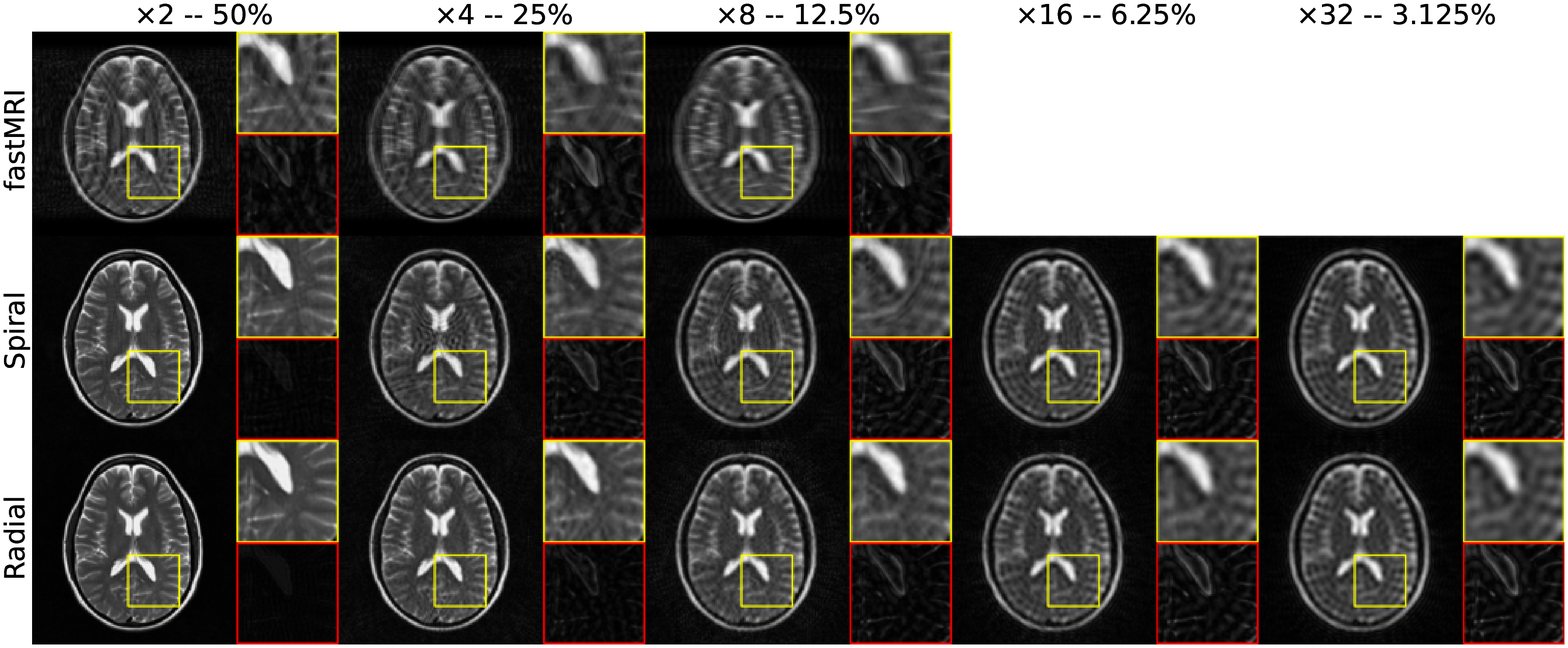}}
\caption{Examples of images obtained using different acceleration factors and \textit{k}-space undersampling masks. The red crop shows the absolute difference between the ground truth and the undersampled images.}
\label{sampling_ex}
\end{center}
\end{figure}

\newpage

\section*{D. Evaluation metrics}

To measure the quality of the images, generated using our acceleration methods, we computed the following metrics:

\textbf{The Mean Square Error (MSE)} is defined by:
$$MSE(x, y)=\frac{1}{MN} \sum_{m=1}^{M} \sum_{n=1}^{N}\bigg(x(m, n)-y(m, n)\bigg)^{2}$$

\textbf{The Peak Signal-to-Noise Ratio (PSNR)} is the ratio between the maximum possible value and the value of the difference between the source and the corrupted image. 
$$PSNR=10 \log _{10}\left(\frac{\mathrm{MAX}_{1}^{2}}{\mathrm{MSE}}\right) =20 \log _{10}\left(\frac{\mathrm{MAX}_{1}}{\sqrt{\mathrm{MSE}}}\right),$$ 
where ${MAX}_{1}$ is the maximum possible value of the image. \\

\textbf{The Structural Similarity Index Measure (SSIM)} is defined by:
$${SSIM}(\mathrm{x}, \mathrm{y})=\frac{\left(2 \mu_{x} \mu_{y}+c_{1}\right)\left(2 \sigma_{x y}+c_{2}\right)}{\left(\mu_{x}^{2}+\mu_{y}^{2}+c_{1}\right)\left(\sigma_{x}^{2}+\sigma_{y}^{2}+c_{2}\right)},$$
where $\mu_{x}$ is the average of $x$;
$\mu_{y}$ is the average of $y$; 
$\sigma_{x}^{2}$ is the variance of $x$;
$\sigma_{y}^{2}$ is the variance of $y$; 
$\sigma_{xy}^{2}$ is the covariance of $x$, $y$;
$c_{1}=\left(k_{1} L\right)^{2},
c_{2}=\left(k_{2} L\right)^{2}$ are constants used for computational stabiltiy; $L$ is the total intensity range; we set $k_1=0.01$ and $k_2=0.03$, as indicated in \cite{ssim}.


\section*{E. Combined models with extreme acceleration factors}
We present here the detailed results obtained with our different models at all acceleration factors.

\begin{figure}[ht]
\begin{center}

\begin{minipage}[h]{0.48\linewidth}

\begin{table}[H]
\scalebox{0.56}{
\begin{tabular}{|p{35pt}|p{76pt}|p{63pt}|p{63pt}|p{63pt}|}
\hline
Mask                       & Model        & MSE, $\times10^{-4}$ & PSNR            & SSIM, $\times 10^{2}$  \\ 
\hline\hline
\multirow{3}{4em}{fastmri} & SRGAN        & 12.98 $\pm$ 3.79  &  29.04 $\pm$ 1.22  &  94.84 $\pm$ 2.55 \\ 
                           & SRGAN+U-Net   & 8.81  $\pm$ 3.08  &  30.80 $\pm$ 1.44  &  95.97 $\pm$ 2.17 \\ 
                           & U-Net+SRGAN   & 9.27  $\pm$ 3.28  &  30.56 $\pm$ 1.41  &  95.69 $\pm$ 2.69 \\
\hline
\multirow{3}{4em}{spiral}  & SRGAN        & 7.66  $\pm$ 2.94  &  31.47 $\pm$ 1.66  &  96.28 $\pm$ 2.26 \\
                           & SRGAN+U-Net   & 7.02  $\pm$ 2.84  &  31.88 $\pm$ 1.73  &  96.51 $\pm$ 2.06 \\ 
                           & U-Net+SRGAN   & 7.09  $\pm$ 2.82  &  31.81 $\pm$ 1.65  &  96.39 $\pm$ 2.37 \\
\hline
\multirow{3}{4em}{radial}  & SRGAN        & 6.25  $\pm$ 2.63  &  32.40 $\pm$ 1.78  &  96.95 $\pm$ 1.95 \\ 
                           & \textbf{SRGAN+U-Net} & \textbf{6.05  $\pm$ 2.64}  &  \textbf{32.58 $\pm$ 1.87}  &  \textbf{97.01 $\pm$ 1.87} \\ 
                           & U-Net+SRGAN   & 6.13  $\pm$ 2.62  &  32.48 $\pm$ 1.75  &  96.92 $\pm$ 2.11 \\
\hline
\end{tabular}
}
\vskip 0.1in
\caption{Low-Resolution ($160 \times 160$) model with $\times2$ undersampling, total acceleration: $\times8$, fraction of \textit{k}-space: $12.5\%$.}

\label{sup:x8_combined_results}
\end{table}

\end{minipage}
\hfill 
\begin{minipage}[h]{0.48\textwidth}%

\begin{table}[H]
\scalebox{0.56}{
\begin{tabular}{|p{35pt}|p{76pt}|p{63pt}|p{63pt}|p{63pt}|}
\hline
Mask                       & Model        & MSE, $\times10^{-4}$ & PSNR            & SSIM, $\times 10^{2}$  \\ 
\hline\hline
\multirow{3}{4em}{fastmri} & SRGAN        & 23.07 $\pm$ 6.30  &  26.53 $\pm$ 1.17  &  92.40 $\pm$ 2.93 \\ 
                           & SRGAN+U-Net   & 15.97 $\pm$ 4.77  &  28.15 $\pm$ 1.26  &  94.00 $\pm$ 2.61 \\ 
                           & U-Net+SRGAN   & 18.55 $\pm$ 5.88  &  27.51 $\pm$ 1.28  &  93.32 $\pm$ 3.27 \\
\hline
\multirow{3}{4em}{spiral}  & SRGAN        & 17.29 $\pm$ 5.05  &  27.79 $\pm$ 1.20  &  93.32 $\pm$ 2.88 \\
                           & SRGAN+U-Net   & 13.61 $\pm$ 4.11  &  28.85 $\pm$ 1.26  &  94.19 $\pm$ 2.58 \\ 
                           & U-Net+SRGAN   & 13.49 $\pm$ 4.18  &  28.89 $\pm$ 1.27  &  94.08 $\pm$ 3.39 \\
\hline
\multirow{3}{4em}{radial}  & SRGAN        & 12.71 $\pm$ 4.07  &  29.18 $\pm$ 1.38  &  94.54 $\pm$ 2.79 \\ 
                           & \textbf{SRGAN+U-Net}   & 11.44 $\pm$ 3.90  &  29.66 $\pm$ 1.46  &  \textbf{94.92 $\pm$ 2.59} \\ 
                           & U-Net+SRGAN   & \textbf{11.39 $\pm$ 3.92}  &  \textbf{29.67 $\pm$ 1.42}  &  94.79 $\pm$ 3.23 \\
\hline
\end{tabular}
}
\vskip 0.1in
\caption{ Low-Resolution ($160 \times 160$) model with $\times4$ undersampling, total acceleration: $\times16$, fraction of \textit{k}-space: $6.25\%$.}
\label{sup:x16_combined_results}
\end{table}

\end{minipage}

\vfill

\begin{minipage}[h]{0.48\linewidth}
\begin{table}[H]
\scalebox{0.56}{
\begin{tabular}{|p{32pt}|p{76pt}|p{63pt}|p{63pt}|p{63pt}|}
\hline
Mask                       & Model        & MSE, $\times10^{-4}$ & PSNR            & SSIM, $\times 10^{2}$  \\ 
\hline\hline
\multirow{3}{4em}{fastmri} & SRGAN        & 49.63 $\pm$ 13.10  &  23.19 $\pm$ 1.14  &  87.19 $\pm$ 3.49 \\ 
                           & SRGAN+U-Net   & 36.99 $\pm$ 9.93   &  24.47 $\pm$ 1.16  &  89.57 $\pm$ 3.10 \\ 
                           & U-Net+SRGAN   & 38.72 $\pm$ 11.65  &  24.30 $\pm$ 1.26  &  89.09 $\pm$ 3.82 \\
\hline
\multirow{3}{4em}{spiral}  & SRGAN        & 30.89 $\pm$ 7.43  &  25.23 $\pm$ 1.04  &  89.86 $\pm$ 3.33 \\
                           & SRGAN+U-Net   & 23.65 $\pm$ 6.13  &  26.40 $\pm$ 1.11  &  91.43 $\pm$ 2.96 \\ 
                           & U-Net+SRGAN   & 24.48 $\pm$ 6.76  &  26.27 $\pm$ 1.17  &  91.16 $\pm$ 3.68 \\
\hline
\multirow{3}{4em}{radial}  & SRGAN        & 25.29 $\pm$ 6.90  &  26.13 $\pm$ 1.16  &  91.34 $\pm$ 3.16 \\ 
                           & \textbf{SRGAN+U-Net}   & \textbf{20.70 $\pm$ 5.91}  &  \textbf{27.01 $\pm$ 1.21}  &  \textbf{92.40 $\pm$ 2.96} \\ 
                           & U-Net+SRGAN   & 20.88 $\pm$ 6.09  &  26.98 $\pm$ 1.22  &  92.13 $\pm$ 3.62 \\
\hline
\end{tabular}
}
\vskip 0.1in
\caption{Low-Resolution ($160 \times 160$) model with $\times2$ undersampling, total acceleration: $\times32$, fraction of \textit{k}-space: $3.125\%$.}
\label{x32_combined_results}
\end{table}

\end{minipage}
\hfill 
\begin{minipage}[h]{0.48\textwidth}

\begin{table}[H]
\raisebox{-1.5cm}{
\scalebox{0.56}{
\begin{tabular}{|p{32pt}|p{76pt}|p{63pt}|p{63pt}|p{63pt}|}
\hline
Mask                       & Model        &  MSE, $\times10^{-4}$ & PSNR            & SSIM, $\times 10^{2}$  \\ 
\hline\hline
\multirow{3}{4em}{spiral}  & SRGAN        & 46.46 $\pm$ 10.66  &  23.44 $\pm$ 1.01  &  86.80 $\pm$ 3.29 \\
                           & SRGAN+U-Net   & 38.09 $\pm$ 9.34   &  24.32 $\pm$ 1.09  &  88.42 $\pm$ 3.16 \\ 
                           & U-Net+SRGAN   & 38.76 $\pm$ 10.41  &  24.27 $\pm$ 1.16  &  88.32 $\pm$ 3.78 \\
\hline
\multirow{3}{4em}{radial}  & SRGAN        & 42.97 $\pm$ 10.13  &  23.79 $\pm$ 1.01  &  87.71 $\pm$ 3.39 \\ 
                           & \textbf{SRGAN+U-Net}   & \textbf{35.20 $\pm$ 8.92}  &  \textbf{24.67 $\pm$ 1.08}  &  \textbf{89.24 $\pm$ 3.19} \\ 
                           & U-Net+SRGAN   & 35.64 $\pm$ 9.38  &  24.63 $\pm$ 1.13  &  89.06 $\pm$ 3.73 \\
\hline
\end{tabular}
}
}
\vskip 0.1in
\caption{Low-Resolution ($160 \times 160$) model with $\times2$ undersampling, total acceleration: $\times64$, fraction of \textit{k}-space: $1.5625\%$.}
\label{x64_combined_results}
\end{table}

\end{minipage}
\end{center}
\end{figure}

\begin{table}[H]
\begin{center}

\begin{small}
\begin{sc}
\scalebox{0.77}{
\begin{tabular}{|p{37pt}|p{105pt}|p{63pt}|p{63pt}|p{63pt}|}
\hline
Acc. /   & \multirow{3}{4em}{Model} & \multirow{3}{8em}{MSE, $\times10^{-4}$}  & \multirow{3}{4em}{PSNR} & \multirow{3}{8em}{SSIM, $\times 10^{2}$}   \\ 
frac. of &&&& \\
\textit{k}-space &&&& \\
\hline\hline

\multirow{2}{4em}{$\times4$ \\ 25\%}    & Pix2pix radial        & 6.98 $\pm$ 2.45       &  31.80 $\pm$ 1.42  &  95.97 $\pm$ 3.12 \\
                                        & \textbf{SRGAN (SR)}   & \textbf{3.03  $\pm$ 1.61}   &  \textbf{35.72 $\pm$ 2.12}  &  \textbf{98.64 $\pm$ 0.93} \\      
\hline
\multirow{2}{4em}{$\times8$ \\ 12.5\%}  & Pix2pix radial        & 13.98 $\pm$ 4.27  &  28.72 $\pm$ 1.22  &  93.41 $\pm$ 3.96 \\
                                        & \textbf{SRGAN+U-Net radial}   & \textbf{6.05  $\pm$ 2.64}  & \textbf{32.58 $\pm$ 1.87}  &  \textbf{97.01 $\pm$ 1.87} \\ 
\hline
\multirow{3}{4em}{$\times16$ \\ 6.25\%} & Pix2pix radial        & 26.14 $\pm$ 7.44  &  25.99 $\pm$ 1.18  &  90.55 $\pm$ 4.20 \\
                                        & \textbf{SRGAN (SR)}   & 11.69 $\pm$ 4.21   &  29.58 $\pm$ 1.49  &  \textbf{95.63 $\pm$ 2.35} \\ 
                                        & U-Net+SRGAN radial           & \textbf{11.39 $\pm$ 3.92} &  \textbf{29.67 $\pm$ 1.42}  &  94.79 $\pm$ 3.23 \\
\hline
\multirow{2}{4em}{$\times32$ \\ 3.125\%} & Pix2pix radial       & 54.83 $\pm$ 16.45  &  22.87 $\pm$ 1.12  &  86.39 $\pm$ 5.21 \\ 
                                         & \textbf{SRGAN+U-Net radial}  & \textbf{20.70 $\pm$ 5.91}  &  \textbf{27.01 $\pm$ 1.21}  &  \textbf{92.40 $\pm$ 2.96} \\ 
\hline
\multirow{2}{4em}{$\times64$ \\ 1.5625\%} & SRGAN (SR)    & \textbf{28.76 $\pm$ 9.67}   &  24.37 $\pm$ 1.13  &  88.49 $\pm$ 3.32 \\ 
                                          & \textbf{SRGAN+U-Net radial}   & 35.20 $\pm$ 8.92   &  \textbf{24.67 $\pm$ 1.08}  &  \textbf{89.24 $\pm$ 3.19} \\ 
\hline
\end{tabular}
}
\end{sc}
\end{small}

\end{center}
\caption{Comparison of the best models selected based on the SSIM metric for the brain dataset.}
\label{models_comp_brain}
\end{table}


\begin{table}[H]
\begin{center}

\begin{small}
\begin{sc}
\scalebox{0.77}{
\begin{tabular}{|p{37pt}|p{105pt}|p{63pt}|p{63pt}|p{63pt}|}
\hline
Acc. /   & \multirow{3}{4em}{Model} & \multirow{3}{8em}{MSE, $\times10^{-4}$}  & \multirow{3}{4em}{PSNR} & \multirow{3}{8em}{SSIM, $\times 10^{2}$}   \\ 
frac. of &&&& \\
\textit{k}-space &&&& \\
\hline\hline

\multirow{2}{4em}{$\times4$ \\ 25\%}    & Pix2pix radial        & 27.95 $\pm$ 19.64       &  26.23 $\pm$ 2.79  &  86.69 $\pm$ 5.83 \\
                                        & \textbf{SRGAN (SR)}   & \textbf{8.84 $\pm$ 9.04}   &  \textbf{32.77 $\pm$ 4.89}  &  \textbf{94.31 $\pm$ 3.65} \\      
\hline
\multirow{2}{4em}{$\times8$ \\ 12.5\%}  & Pix2pix radial        & 42.26 $\pm$ 29.35  &   24.54 $\pm$ 2.99  &  80.97 $\pm$ 8.20 \\
                                        & \textbf{SRGAN+U-Net radial}   & \textbf{21.58 $\pm$ 18.77}  & \textbf{27.62 $\pm$ 3.23}  &  \textbf{88.74 $\pm$ 5.35} \\ 
\hline
\multirow{3}{4em}{$\times16$ \\ 6.25\%} & Pix2pix radial        & 58.09 $\pm$ 40.44  &  23.22 $\pm$ 3.02  &  74.87 $\pm$ 10.44 \\
                                        & \textbf{SRGAN (SR)}   & \textbf{26.02 $\pm$ 21.85}   &  \textbf{27.53 $\pm$ 4.23}  &  \textbf{85.04 $\pm$ 7.92} \\ 
                                        & U-Net+SRGAN radial     & 28.40 $\pm$ 22.85 &  26.53 $\pm$ 3.31  &  83.98 $\pm$ 8.42 \\
\hline
\multirow{2}{4em}{$\times32$ \\ 3.125\%} & Pix2pix radial       & 77.72 $\pm$ 52.45  &  21.86 $\pm$ 2.84  &  70.65 $\pm$ 11.57 \\ 
                                         & \textbf{SRGAN+U-Net radial}  & \textbf{42.48 $\pm$ 35.12}  &  \textbf{24.97 $\pm$ 3.56}  &  \textbf{80.67 $\pm$ 8.90} \\ 
\hline
\multirow{2}{4em}{$\times64$ \\ 1.5625\%} & \textbf{SRGAN (SR)}    & \textbf{45.71 $\pm$ 34.51}   &  \textbf{24.82 $\pm$ 3.88}  &  \textbf{76.95 $\pm$ 10.65} \\ 
                                          & SRGAN+U-Net radial   & 56.72 $\pm$ 41.34   &  23.44 $\pm$ 3.19  &  76.24 $\pm$ 10.11 \\ 
\hline
\end{tabular}
}
\end{sc}
\end{small}

\end{center}
\caption{Comparison of the best models, selected based on the SSIM metric for the knee dataset.}
\label{models_comp_knee}
\end{table}

\newpage
\section*{F. Labeling tool for user study}

We used the community version of the open-source labeling tool Label Studio \cite{label-studio} to perform the user study described in the main text, Section
5.
We adjusted the user interface to compare the accelerated (left) and the ground truth (right) images based on tree IQ criteria on a 4-point scale. 

\begin{figure}[H]
    \centering
    \includegraphics[width=\textwidth]{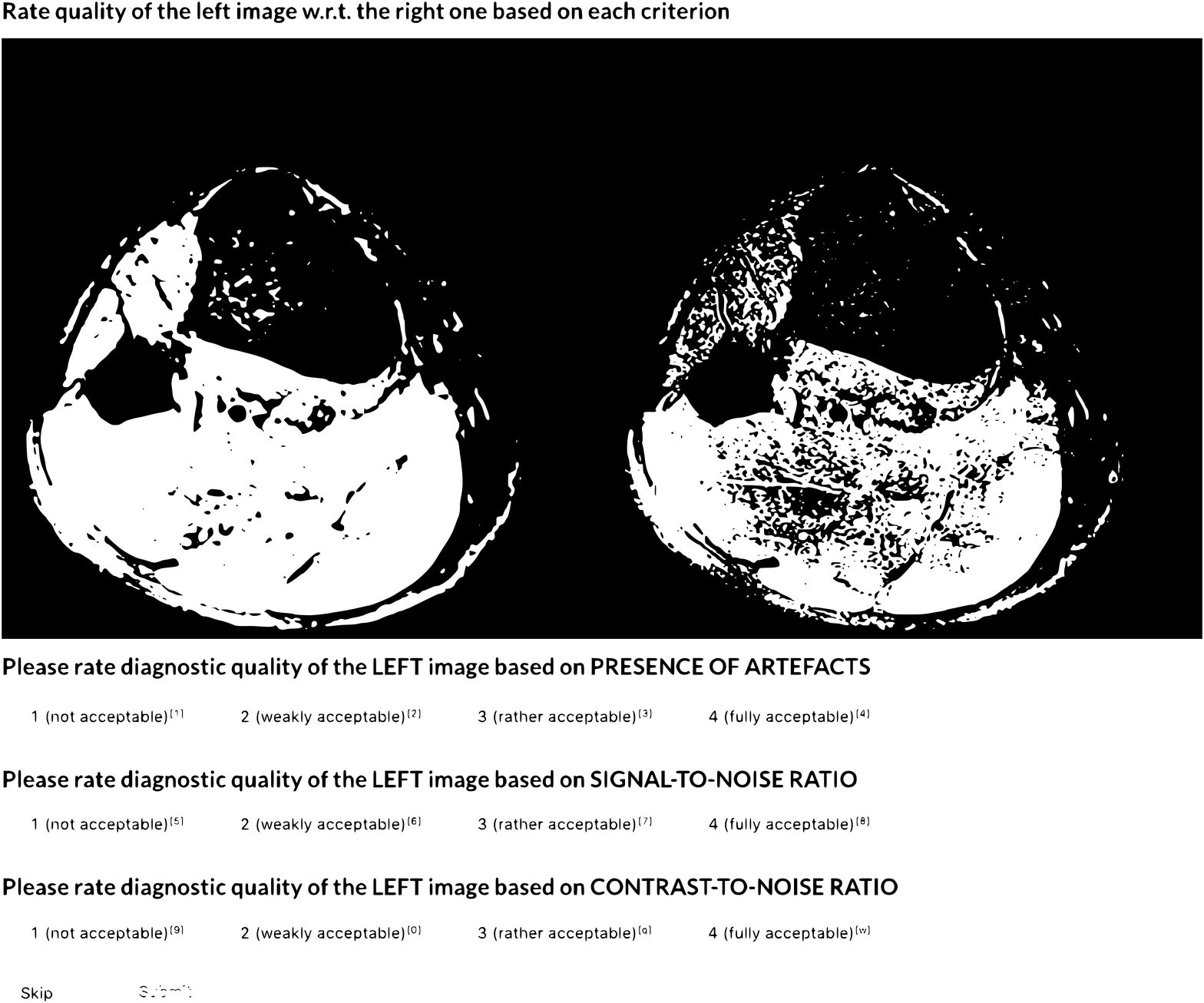}
    \caption{User interface of the tool for the user study. It was used to survey five radiologists about the diagnostic value of the reconstructions.}
    \label{fig:user-interface}
\end{figure}

An example of the user interface is presented in figure \ref{fig:user-interface}.
Experts were asked to rate the quality of the left (accelerated) image compared to the right one (ground truth).
The main element of the interface is the panel section with two images for the radiologists to observe side-by-side.
Users can zoom in and zoom out the images using a mouse scroll wheel to conveniently examine the regions of interest.
While performing the examination, the users provide their answers to the questions below, using mouse or keyboard shortcuts.
Once the evaluation is finished, users press submit button and proceed to another pair of images. 
Users can also skip a pair of images in case if they face difficulties with scoring a particular case.
The examination is finished either when the user decides to stop or when there are no pairs of images left in the test subset.

\section*{G. Experimental Setup}

\subsection*{Specification of dependencies}

All models were implemented using \texttt{Python 3.8} and \texttt{PyTorch 1.4}. Training models and all the experiments were conducted using Nvidia Tesla V100 with 32 GB RAM.

Adam optimizer is used to train the models, with the initial learning rate being equal to $10^{-4}$. The size of a batch is 4, and the number of epochs is 100. The combined model contains $29.1$M trainable parameters.

\subsection*{Complexity analysis}

Training time of the combined models (SRGAN+U-Net, U-Net+SRGAN) are about 60 hours. An average inference time for 100000 slices shown in Table \ref{time_complexity}.

Training the combined model with the above parameters requires about 11GB of GPU memory. The inference of the combined model requires 2.2GB of GPU memory.

\begin{table}[H]
\begin{center}

\begin{small}
\begin{sc}
\begin{tabular}{|p{80pt}|p{50pt}|p{100pt}|}
\hline
Model & Time, ms & \# of Parameters, M \\

\hline\hline
Pix2pix   & 9.6 &   26.1\\    
\hline

SRGAN(SR) $\times 4$   & 7.2 &  5.7\\      
SRGAN(SR) $\times 8$   & 6.1 &  5.9 \\        
SRGAN(SR) $\times 64$   & 5.8 &  6.2 \\        
\hline
U-Net+SRGAN   & 10.7  & 29.1 \\     
SRGAN+U-Net   & 16.9 &  29.1 \\      
                                         
\hline
\end{tabular}
\end{sc}
\end{small}

\end{center}
\caption{An average inference time of the models per slice in milliseconds and the number of trainable parameters.}
\label{time_complexity}
\end{table}